\documentclass[journal]{IEEEtran}

\usepackage{amsmath,amsfonts}
\usepackage{algorithmic}
\usepackage{algorithm}
\usepackage{array}
\usepackage{textcomp}
\usepackage{stfloats}
\usepackage{url}
\usepackage{verbatim}
\usepackage{graphicx}   
\usepackage{cite}
\usepackage{xcolor}
\usepackage{subcaption}

\usepackage{tabularx}
\usepackage{multirow}
\usepackage{booktabs}
\usepackage{makecell}

\usepackage{rotating}
\usepackage{threeparttable}
\usepackage[colorlinks=true, linkcolor=blue, urlcolor=blue, citecolor=blue]{hyperref}

\usepackage{cleveref}
\Crefformat{figure}{#2Fig.~#1#3}
\Crefmultiformat{figure}{Figs.~#2#1#3}{ and~#2#1#3}{, #2#1#3}{ and~#2#1#3}
\begin{document}

\title{\fontsize{23.3pt}{30pt}\selectfont CycleRL: Sim-to-Real Deep Reinforcement Learning for Robust Autonomous Bicycle Control}

\author{Gelu Liu, Teng Wang, Zhijie Wu, Junliang Wu, Songyuan Li, Xiangwei Zhu
        % <-this % stops a space
\thanks{The paper was supported by the National Natural Science Foundation of China under Grant Nos. T2350005 and T2541041, and Shenzhen Science and Technology Program under Grant No. ZDSYS20210623091807023. \textit{(Gelu Liu and Teng Wang are co-first authors.) (Corresponding author: Xiangwei Zhu.)}}

\thanks{Gelu Liu, Teng Wang, Zhijie Wu, Junliang Wu, Songyuan Li, Xiangwei Zhu are with the School of Electronics and Communication Engineering and Shenzhen Key Laboratory of Navigation and Communication Integration, Sun Yat-sen University, Shenzhen 518107, China (e-mail: liuglu@mail2.sysu.edu.cn; wangt556@mail.sysu.edu.cn; wuzhj39@mail2.sysu.edu.cn; wujliang9@mail2.sysu.edu.cn; lisy287@mail.sy\\su.edu.cn; zhuxw666@mail.sysu.edu.cn).}

}

\markboth{IEEE Robotics and Automation Letters}%
{Shell \MakeLowercase{\textit{et al.}}: A Sample Article Using IEEEtran.cls for IEEE Journals}

\IEEEpubid{0000--0000/00\$00.00~\copyright~2026 IEEE}

\maketitle

\begin{abstract}

    Autonomous bicycles offer a promising agile solution for urban mobility and last-mile logistics. However, conventional control strategies often struggle with underactuated nonlinear dynamics, suffering from sensitivity to model mismatches and limited adaptability to real-world uncertainties. To address this, we develop CycleRL, a comprehensive sim-to-real framework for robust autonomous bicycle control. Our approach establishes a direct perception-to-action mapping within the high-fidelity NVIDIA Isaac Sim environment, leveraging Proximal Policy Optimization (PPO) to optimize the control policy. The framework features a composite reward function tailored for concurrent balance maintenance, velocity tracking, and steering control. Crucially, systematic domain randomization is employed to reduce the reliance on precise system modeling, bridge the simulation-to-reality gap and facilitate direct transfer. In simulation, CycleRL achieves promising performance, including a 99.90\% balance success rate, a heading tracking error of 1.15°, and a velocity tracking error of 0.18 m/s. These quantitative results, coupled with successful hardware deployment, validate DRL as an effective paradigm for autonomous bicycle control, offering superior adaptability over traditional methods. Video demonstrations are available at \href{https://cpnt-lab.github.io/CycleRL/}{https://cpnt-lab.github.io/CycleRL/}.
    
    % Autonomous bicycle control represents a challenging problem in underactuated nonlinear systems, requiring precise lateral balance and steering coordination under dynamic conditions. Traditional model-based approaches suffer from sensitivity to parameter uncertainties and computational complexity, limiting their real-world applicability. This paper presents a comprehensive deep reinforcement learning framework for robust lateral neural control of autonomous bicycles. Our approach eliminates the need for explicit dynamic modeling by training end-to-end neural policies through Proximal Policy Optimization (PPO) in high-fidelity NVIDIA Isaac Sim environments. The framework incorporates a carefully designed multi-objective reward function that simultaneously optimizes balance maintenance, velocity tracking, and steering precision. To address the simulation-to-reality gap, we implement systematic domain randomization across vehicle dynamics, environmental conditions, and sensor characteristics. Comprehensive experimental validation demonstrates successful transfer to hardware implementation with robust performance across diverse operational conditions including varying velocities, terrain types, and payload configurations. These results establish deep reinforcement learning as a viable paradigm for practical autonomous bicycle control, offering considerable adaptability and robustness compared to traditional model-based approaches.

\end{abstract}

\begin{IEEEkeywords}
Deep reinforcement learning, autonomous bicycle control, sim-to-real transfer, domain randomization, underactuated systems, lateral control.
\end{IEEEkeywords}

\section{Introduction}
\label{sec:intro}

The autonomous control of two-wheeled vehicles is a classic challenge in robotics, involving stabilizing inherently unstable dynamics while executing steering maneuvers~\cite{yeh2024robust}. Such capability is critical for emerging applications such as autonomous cycling, micro-logistics, and intelligent mobility solutions, which demand robust stability and precise lateral control in complex urban environments~\cite{kuutti2020survey}. Consequently, autonomous bicycle control has become a key testbed for advancing control methodologies and embodied intelligence.

\begin{figure}[htbp]
    \centering
    \includegraphics[width=0.95\linewidth]{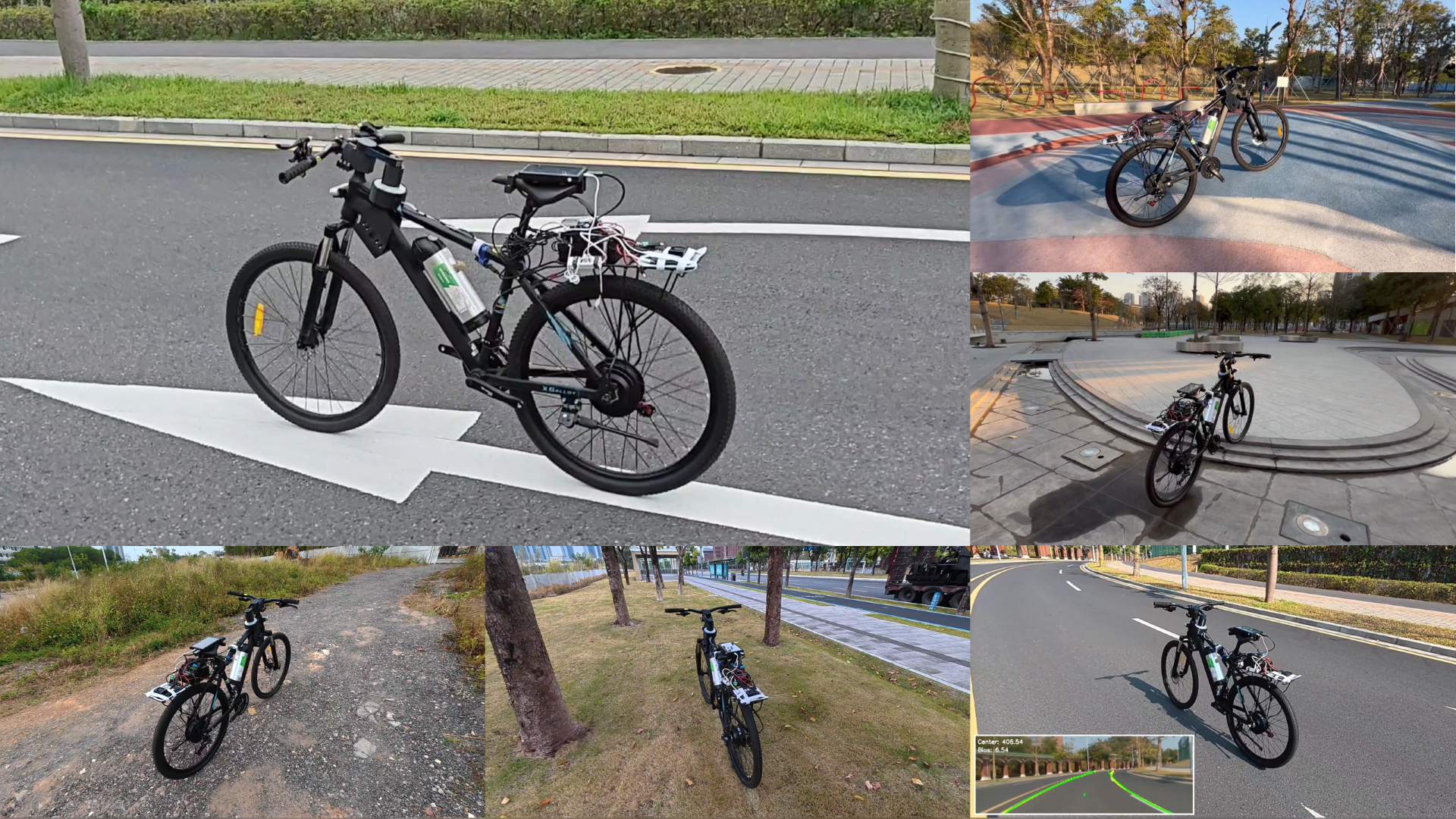}
    \caption{Real-world deployment of the proposed autonomous bicycle control framework across different conditions.}
    \label{fig:intro}
\end{figure}

Two primary paradigms exist for self-balancing systems: reaction wheels and steering control. While reaction-wheel-based approaches offer static balancing capabilities~\cite{huang2025stability, zhu2024online}, they are often limited by high energy consumption, increased mechanical complexity, and payload constraints. In contrast, steering control presents a more pragmatic and efficient alternative that aligns with natural bicycle dynamics~\cite{deng2018towards, baquero2019robust}. Its simplified architecture, lower energy requirements, and superior load-carrying capacity make it the preferred choice for practical, real-world applications.

\IEEEpubidadjcol

Historically, steering control has been implemented via classical methods, ranging from PID controllers~\cite{yu2018steering, pei2019towards}, LQR~\cite{huang2025stability}, MPC~\cite{bellegarda2021dynamic}, LPV~\cite{yeh2024robust}, to fuzzy logic~\cite{shafiekhani2015design}. However, these approaches face significant drawbacks. PID controllers demand extensive manual tuning, while model-based methods, such as MPC, are computationally intensive and highly sensitive to inaccuracies in system parameters. This reliance on precise models often results in controllers that are brittle to real-world variations, compromising their reliability.

These limitations have motivated the shift toward Deep Reinforcement Learning (DRL), a powerful paradigm for solving complex sequential decision-making problems with high-dimensional inputs and nonlinear dynamics~\cite{folkers2019controlling}. By leveraging high-fidelity simulators, such as Isaac Gym~\cite{makoviychuk2021isaac} and MuJoCo~\cite{todorov2012mujoco}, DRL enables agents to learn optimal control strategies through autonomous interaction, thereby reducing the reliance on highly accurate analytical models. This paradigm establishes a direct mapping from sensory observations to actuator commands, which can reduce system latency and optimize the control pipeline, offering superior robustness and adaptability for dynamic environments~\cite{li2019reinforcement}.

% In this work, we present a DRL framework for end-to-end autonomous bicycle control. We leverage Proximal Policy Optimization (PPO)~\cite{schulman2017proximal} to train the control policiy within Isaac Sim environment, incorporating domain randomization to ensure robust sim-to-real transfer. The resulting system is validated on a custom-built bicycle platform, demonstrating effective end-to-end control without explicit dynamic modeling, as shown in \Cref{fig:intro}. The final system achieves impressive real-world performance metrics, including a balance success rate of 99.90\%, an algorithmic latency of 1 ms, and a sustained balance duration exceeding 30 minutes, highlighting the practical effectiveness of our approach.

In this work, we present CycleRL, a model-free deep reinforcement learning (DRL) framework for autonomous bicycle control that reduces the reliance on accurate analytical dynamics models. Trained within Isaac Sim, our approach designed a tailored composite reward function to simultaneously optimize balance, velocity, and steering. To bridge the reality gap, we implement a comprehensive domain randomization strategy that facilitates reliable sim-to-real transfer. Extensive evaluations demonstrate that CycleRL achieves a 99.90\% simulated balance success rate—outperforming classical baselines—and exhibits exceptional real-world robustness on a custom physical platform against diverse conditions, as shown in \Cref{fig:intro}.

The remainder of this paper is structured as follows: \Cref{sec:related} reviews related work in autonomous control and DRL. \Cref{sec:method} details our methodological framework. \Cref{sec:eval} presents our experimental evaluations, and \Cref{sec:conclusion} concludes the paper. Our primary contributions are:

\begin{itemize}
    \item A comprehensive system integration of a model-free Deep Reinforcement Learning (DRL) framework for autonomous bicycle control, which alleviates the dependence on accurate analytical dynamics models required by conventional methods.

    \item A carefully engineered composite reward function coupled with a multi-layered domain randomization strategy, enabling favorable simulated control performance and reliable direct sim-to-real transferability.

    \item A successful sim-to-real transfer on a custom-built bicycle platform, with rigorous real-world validation demonstrating robust and promising balance performance under various physical conditions.
\end{itemize}

\section{Related Work}
\label{sec:related}
\subsection{Traditional Control Methods for Two-Wheeled Vehicles}

Traditional model-based methodologies, including Propor-tional-Integral-Derivative (PID)~\cite{yu2018steering}, Linear-Quadratic Regulator (LQR)~\cite{huang2025stability}, and Model Predictive Control (MPC)~\cite{bellegarda2021dynamic}, have been extensively explored for two-wheeled vehicle control. While these methods improve responsiveness and constraint handling, they fundamentally struggle with parameter sensitivity and lack robustness against unmodeled dynamics.

Alternatively, knowledge-based approaches, such as fuzzy logic, manage system nonlinearities without requiring a precise mathematical model. Fuzzy controllers have been successfully designed to achieve robust balance and steering control under different conditions~\cite{shafiekhani2015design, hwang2009fuzzy}. However, their performance is highly dependent on extensive expert knowledge and manual tuning, which can result in suboptimal control strategies.

Despite their foundational contributions, these traditional methods generally share common limitations: high sensitivity to model inaccuracies, significant computational demand, and limited adaptability across diverse conditions. These challenges motivate the exploration of learning-based approaches that can derive robust control policies directly from environmental interactions.

% 【旧】The control of two-wheeled vehicles has been extensively addressed by traditional control methodologies. Model-based approaches such as Proportional-Integral-Derivative (PID), Linear-Quadratic Regulator (LQR), and Model Predictive Control (MPC) have been widely explored. While LQR controllers have shown to improve responsiveness and efficiency~\cite{huang2017balancing, huang2025stability, li2025variable} and MPC offers superior constraint handling for trajectory tracking~\cite{bellegarda2021dynamic}, these methods fundamentally rely on accurate system models and incur significant computational overhead.

% Alternatively, knowledge-based approaches like fuzzy logic manage nonlinearities without precise mathematical models, achieving robust balance and steering~\cite{shafiekhani2015design, hwang2009fuzzy}. However, their reliance on extensive expert knowledge and manual tuning often yields suboptimal control strategies. 

% Overall, the high sensitivity to model inaccuracies, significant computational demand, and limited adaptability of traditional methods motivate the exploration of learning-based approaches that derive robust policies directly from interaction.

\subsection{Learning-Based Applications in Bicycle Control}

Learning-based approaches, notably imitation learning (IL) and deep reinforcement learning (DRL), have emerged as transformative paradigms in embodied control~\cite{kuutti2020survey}. Advances in simulators, e.g., NVIDIA Isaac Gym~\cite{makoviychuk2021isaac} and MuJoCo~\cite{todorov2012mujoco}, have enabled massive parallel experience collection and efficient policy training, lowering the barrier for deploying neural controllers in nonlinear and underactuated systems.

IL has shown promise in bicycle motion control, primarily by leveraging expert demonstrations to achieve rapid convergence in balancing and trajectory tracking. For instance, Weissmann et al. used conditional IL for vision-based bicycle trajectory prediction, achieving favorable short-term accuracy~\cite{weissmann2021empirical}. However, IL policies tend to generalize poorly beyond the training distribution and often require high-quality demonstrations, which are hard to acquire in rare scenarios, making IL less suitable for bicycle control in the real world.

In contrast, DRL enables learning control policies through trial-and-error interactions without relying on expert data, making it highly suitable for dynamically coupled vehicle control~\cite{liu2023reinforcement}. For instance, DRL has been employed to integrate multiple control objectives such as path tracking and stability in autonomous scooter driving scenarios~\cite{baltes2023deep}, and to achieve robust navigation for vehicles traversing rough and varied terrains~\cite{wiberg2022control}. Focusing on the particularly challenging domain of two-wheeled vehicles, Guo et al. developed an RL algorithm based on the nonaffine nonlinear dynamics of a bicycle, successfully demonstrating self-balancing control with preliminary physical validation~\cite{guo2024combined}.

Despite significant progress in simulation, achieving robust sim-to-real transfer for bicycles controlled solely through steering (i.e., without reaction wheels) remains a major engineering challenge. This difficulty stems from: (1) the system's inherent instability, particularly at low speeds, and (2) its extreme sensitivity to minute physical discrepancies between the simulated and real-world models. This represents a substantial gap in advancing autonomous mobility solutions.

\subsection{Simulation-to-Reality Transfer Technologies}

A central challenge in deploying learning-based controllers is bridging the ``simulation-to-reality'' (sim-to-real) gap, where policies trained in simulation exhibit a significant performance degradation on physical systems~\cite{hofer2021sim2real}. This performance drop is attributable to the discrepancies between the simulated model and real-world dynamics, such as parametric inaccuracies, unmodeled stochastic dynamics, actuator discrepancies, sensor noise, and external disturbances.

Key strategies have been developed to bridge this gap. Domain randomization enhances policy generalization by training across a wide distribution of simulation parameters, a technique proven effective for transferring vision-based policies to real robots~\cite{tobin2017domain}. Other prominent methods include training ensembles of dynamics models to capture uncertainty, using adversarial training frameworks such as Robust Adversarial Reinforcement Learning (RARL) to prepare for worst-case scenarios~\cite{10702590}, and employing domain adaptation to align simulation and real-world distributions~\cite{truong2021bi}.

However, the application of these sim-to-real techniques to the complex dynamics of bicycle control remains largely unexplored. This presents a significant opportunity to advance the real-world deployment of autonomous two-wheeled vehicles, given their unique stability challenges and high sensitivity to environmental factors.

% A central challenge in deploying learning-based controllers is the "sim-to-real" gap, where simulation-trained policies degrade on physical systems due to parametric inaccuracies, unmodeled stochastic dynamics, and sensor noise~\cite{hofer2021sim2real}.

% Various strategies address this gap. Domain randomization enhances generalization by training across broad simulation parameter distributions~\cite{tobin2017domain}. Other prominent methods include dynamics model ensembles for uncertainty capture, Robust Adversarial RL (RARL) for worst-case scenario preparation~\cite{10702590}, and domain adaptation to align sim-and-real distributions~\cite{truong2021bi}.

% However, applying these sim-to-real techniques to the complex, highly sensitive dynamics of bicycle control remains largely unexplored. This presents a significant opportunity to advance the real-world deployment of autonomous two-wheeled vehicles.

\section{Methodology}
\label{sec:method}
\subsection{Nonlinear Dynamics and Uncertainty Analysis}

\subsubsection{Two-Wheeled Vehicle Dynamics}

Two-wheeled vehicles represent a canonical class of underactuated nonholonomic systems characterized by intricate coupling between rotational and translational motion that poses a significant challenge to conventional linear control synthesis.

\begin{figure}[htbp]
    \centering
    \includegraphics[width=0.75\linewidth]{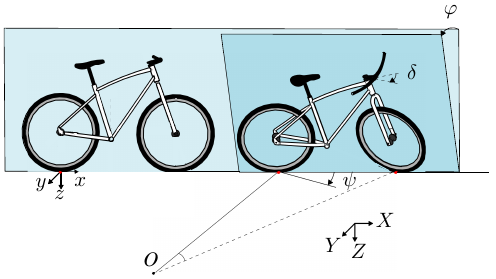}
    \caption{Sketch of the two-wheeled vehicle~\cite{persson2021trajectory}.}
    \label{fig:bicycle-sketch}
\end{figure}

Consider the configuration space parameterized by generalized coordinates $q = [\varphi, \delta, x, y, \psi]^T$, where $\varphi$ represents the roll angle, $\delta$ denotes the steering angle, $(x,y)$ defines the planar position, and $\psi$ is the yaw angle, as shown in \Cref{fig:bicycle-sketch}. The complete nonlinear dynamics can be expressed as:
\begin{equation}
    M_\theta(q)\ddot{q} + C_\theta(q,\dot{q})\dot{q} + G_\theta(q) = B(q)u + \tau_{\text{ext}}(t)
\end{equation}
where $M_\theta(q) \in \mathbb{R}^{5 \times 5}$ is the configuration-dependent inertia matrix, $C_\theta(q,\dot{q})$ captures velocity-dependent Coriolis and gyroscopic effects, $G_\theta(q)$ represents gravitational and centrifugal forces, $B(q)$ is the input distribution matrix, and $\tau_{\text{ext}}(t)$ denotes external disturbances~\cite{meijaard2007linearized}.

\vspace{-1em}

\subsubsection{Stochastic Dynamics Formulation}

To capture real-world uncertainties, we model the bicycle's system evolution via a stochastic differential equation (SDE):
\begin{equation}
    \label{eq:sde}
    dq = f_\theta(q, u) dt + g_\theta(q) dW_t + h_\theta(q) d\xi_t
\end{equation}
where $\theta$ denotes uncertain physical parameters (e.g., mass distribution, friction). Here, $f_\theta(q, u)$ governs the nominal nonlinear dynamics, while environmental stochasticity is captured by two terms: $g_\theta(q) dW_t$ models continuous disturbances (e.g., terrain vibrations) via a standard Wiener process $W_t$, and $h_\theta(q) d\xi_t$ models discrete impulsive events (e.g., sudden road steps) via a Poisson jump process $\xi_t$~\cite{yasuda2023stochastic}.

Analytically solving the optimal control problem for Eq. \ref{eq:sde} is mathematically intractable. While traditional approaches like nonlinear MPC can approximate solutions, they typically require stringent simplifying assumptions, exact explicit system identification, or computationally prohibitive online optimization. To bypass these modeling and computational bottlenecks, we adopt a model-free DRL framework, as detailed below.

\vspace{-1em}
\subsection{Reinforcement Learning Framework}

\subsubsection{Markov Decision Process Formulation}

We formulate the control problem as a Markov Decision Process (MDP), defined by the tuple $\mathcal{M} = (\mathcal{S}, \mathcal{A}, \mathcal{P}, \mathcal{R}, \gamma)$. Here, $\mathcal{S} \in \mathbb{R}^n$ and $\mathcal{A} \in \mathbb{R}^m$ denote the continuous state and action spaces, respectively. $\mathcal{P}$ governs the transition dynamics, $\mathcal{R}$ defines the control objectives, and $\gamma$ is the discount factor. Although the physical system is partially observable, we utilize the instantaneous observation vector $o_t$ (comprising IMU and encoder data) directly as the state $s_t$. This approximation is justified by the high control frequency, which mitigates the need for temporal history integration. Consequently, we learn a reactive stochastic policy $\pi_\theta(a_t|s_t)$ parameterized by a standard Multi-Layer Perceptron (MLP), relying on domain randomization to ensure robustness against sensor noise.

\vspace{0.15cm}

\subsubsection{Composite Reward Function Design}

\begin{figure}[htbp]
    \centering
    \includegraphics[width=0.9\linewidth]{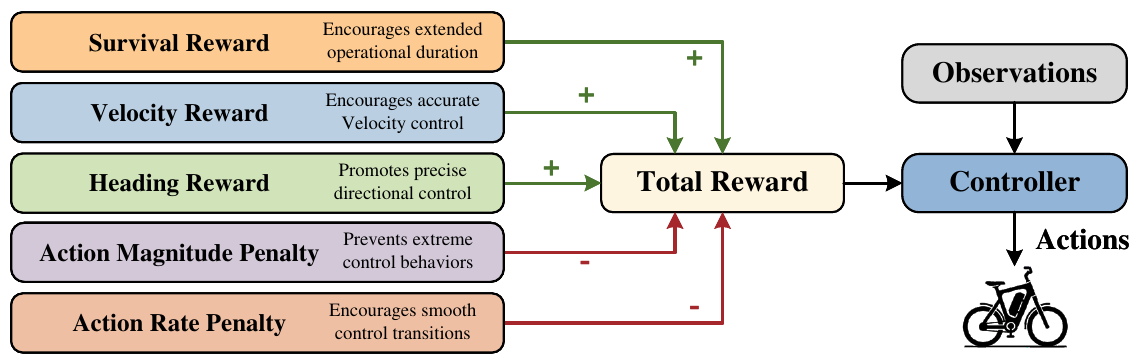}
    \caption{Illustration of the reward function design for balancing and steering control. The total reward aggregates performance incentives (green arrows) and control effort penalties (red arrows) into a unified scalar to guide the policy learning.}
    \label{fig:RL_reward}
\end{figure}

The controller's learning process is guided by a composite reward function designed to balance performance goals with control efficiency (illustrated in \Cref{fig:RL_reward}). By aggregating positive incentives (for stability and tracking) and negative penalties (for action smoothness) into a single scalar signal, we transform the multi-objective requirements into a tractable optimization problem.

% The controller's learning process is guided by a composite reward function designed to balance performance goals with control efficiency. As illustrated in \Cref{fig:RL_reward}, this function serves as a reward shaping mechanism, combining positive incentives for achieving stability and tracking objectives with negative penalties to ensure smooth and economical actions. By aggregating these components into a single scalar signal, we transform the multi-objective requirements into a tractable optimization problem.

The total reward $R_t$ at each timestep is computed as the weighted sum of five key components:
\begin{equation}
    \label{eq:total_reward}
    \begin{aligned}
        R_t = { }&\lambda_{\text{surv}} \cdot r_{\text{surv}} + \lambda_{\text{vel}} \cdot r_{\text{vel}} + \lambda_{\text{head}} \cdot r_{\text{head}} + \\
        &\lambda_{\text{act}} \cdot r_{\text{act}} + \lambda_{\text{rate}} \cdot r_{\text{rate}}
    \end{aligned}
\end{equation}
where the weights are set to $\lambda_{\text{surv}}=1.0$, $\lambda_{\text{vel}}=3.0$, $\lambda_{\text{head}}=5.0$, $\lambda_{\text{act}}=1.0$, and $\lambda_{\text{rate}}=2.0$. These components are organized into two groups: performance-oriented rewards and control effort penalties.

% \paragraph{Performance-Oriented Rewards}
\vspace{0.1cm}
\textbf{a) \textit{Performance-Oriented Rewards}}: These components provide positive reinforcement for successfully executing the primary balancing and tracking control tasks.

\textbf{Survival Reward} ($r_{\text{surv}}$): To encourage long-term stability and faster convergence, a constant positive reward is given for each timestep the bicycle remains operational. This incentivizes the agent to avoid falling:
\begin{equation}
    \label{eq:survival_reward}
    \begin{aligned}
        r_{\text{surv}} = 1
    \end{aligned}
\end{equation}

\textbf{Velocity Tracking Reward} ($r_{\text{vel}}$): This reward encourages the agent to accurately follow the target forward velocity ($v_{\text{cmd}}$). The reward is exponentially proportional to the negative error between the actual and commanded velocities:
\begin{equation}
    \label{eq:velocity_reward}
    \begin{aligned}
        r_{\text{vel}} = \exp(-\alpha |v_{\text{actual}} - v_{\text{cmd}}|)
    \end{aligned}
\end{equation}
where the error sensitivity $\alpha = 0.25$.

\textbf{Heading Tracking Reward} ($r_{\text{head}}$): Similarly, this component rewards the agent for precise directional control by minimizing the heading error between the actual ($\psi_{\text{actual}}$) and desired ($\psi_{\text{desired}}$) heading, where $\beta = 0.1$:
\begin{equation}
    \label{eq:heading_reward}
    \begin{aligned}
        r_{\text{head}} = \exp(-\beta |\psi_{\text{actual}} - \psi_{\text{desired}}|)
    \end{aligned}
\end{equation}

% \paragraph{Control Effort Penalties}
\vspace{0.1cm}
\textbf{b) \textit{Control Effort Penalties}}: These components are negative rewards designed to regularize the control policy, leading to smoother and more efficient actions.

\textbf{Action Magnitude Penalty} ($r_{\text{act}}$): A penalty proportional to the L2 norm of the action ($a_t$) is applied to discourage aggressive policy outputs, which promotes energy efficiency:
\begin{equation}
    \label{eq:action_penalty}
    \begin{aligned}
        r_{\text{act}} = -\|a_t\|_2
    \end{aligned}
\end{equation}

\textbf{Action Rate Penalty} ($r_{\text{rate}}$): To ensure smooth control outputs, a penalty is imposed on the L2 norm of the change in action between consecutive timesteps. This discourages high-frequency oscillations (jerkiness):
\begin{equation}
    \label{eq:rate_penalty}
    \begin{aligned}
        r_{\text{rate}} = -\|a_t - a_{t-1}\|_2
    \end{aligned}
\end{equation}

While the DRL approach avoids deriving explicit analytical controllers, the pipeline still involves important engineering decisions, particularly in reward shaping. The hyperparameters were systematically selected based on a hierarchical control logic and sensitivity analysis (detailed in \Cref{sec:ablation}). Tracking rewards ($\lambda_{\text{vel}}, \lambda_{\text{head}}$) are assigned higher weights than the survival baseline to prevent the agent from converging to a trivial ``stand-still'' policy. Additionally, the steering sensitivity is set with a tighter tolerance ($\beta < \alpha$) due to the critical coupling between lateral deviations and roll instability, while penalty weights are calibrated to constrain control frequencies within the physical actuator bandwidth.

% \paragraph{Episode Termination Criteria}
\vspace{0.1cm}
\textbf{c) \textit{Episode Termination Criteria}}:
Two distinct termination conditions are implemented to balance training efficiency with safety considerations. Timeout termination limits episodes to 64 seconds (3,200 steps at 50 Hz), while safety-based termination immediately ends episodes when $|\varphi_t| > 45^\circ$, providing clear failure signals while preventing continuation in physically unrealistic states.

\vspace{0.15cm}

\subsubsection{Proximal Policy Optimization with Composite Reward}

We employ the Proximal Policy Optimization (PPO)~\cite{schulman2017proximal} algorithm to train the control policy. To handle multiple control objectives, the weighted composite reward $R_t$ is linearly scalarized into a unified signal for computing the Generalized Advantage Estimation (GAE). The policy $\pi_\theta$ is then optimized to maximize the expected cumulative return using the standard clipped surrogate objective:
\begin{equation}
    \label{eq:ppo_clip}
    \begin{aligned}
        \mathcal{L}(\theta) = \mathbb{E}_t\left[\min\left(r_t(\theta)\hat{A}_t, \text{clip}(r_t(\theta), 1-\epsilon, 1+\epsilon)\hat{A}_t\right)\right]
    \end{aligned}
\end{equation}
where $r_t(\theta) = \frac{\pi_\theta(a_t|s_t)}{\pi_{\theta_{\text{old}}}(a_t|s_t)}$ represents the probability ratio be-tween the current and old policies, $\hat{A}_t$ denotes the advantage estimate derived from the composite reward $R_t$, and $\epsilon$ is the clipping threshold that constrains the policy ratio to the range $[1-\epsilon, 1+\epsilon]$, thereby ensuring training stability.

% It is worth noting that the exponential design of the tracking rewards (Eq. 5 and Eq. 6) plays a critical role in this integration. Unlike sparse binary rewards or linear error penalties, the exponential formulation provides continuous, dense gradients even when the agent is close to the target state. This ensures that the advantage estimates $\hat{A}_t$ remain informative throughout the training process, enabling the PPO clipped surrogate objective (Eq. 9) to effectively refine the policy for high-precision control.

\subsection{Domain Randomization Strategy}

Sim-to-real discrepancies—ranging from unmodeled dynamics to unseen states—often cause catastrophic hardware failures~\cite{tobin2017domain}. To bridge this gap, we employ a decoupled hierarchical randomization strategy comprising \textit{Dynamics}, \textit{Initial State}, and \textit{Task Randomization}, as shown in \Cref{tab:actual_domain_randomization}. Tailored to underactuated bicycles, this multi-layered approach prevents overfitting to simulation artifacts and forces the policy to generalize across diverse physical conditions.

\begin{table}[htbp]  
\centering  
\caption{Domain Randomization Variables}  
\label{tab:actual_domain_randomization}  
\begin{tabularx}{0.93\columnwidth}{l X}  
\toprule  
\textbf{Randomization Variables} & \textbf{Range/Distribution} \\
\midrule  
\multicolumn{2}{c}{\textbf{\textit{Dynamics Randomization (Physical Parameters)}}} \\
System Total Mass ($m_{\text{total}}$) & $\mathcal{U}([15.0, 45.0])$ kg \\
Center of Mass Height ($h_{\text{CoM}}$) & $\mathcal{U}([0.50, 0.80])$ m \\
Tire Friction Coefficient ($\mu$) & $\mathcal{U}([0.5, 1.2])$ \\
Actuator Force Gain & $\mathcal{U}([0.9, 1.1])$ \\
Observation Noise & \makecell[l]{$\mathcal{N}(0,\ (\rho \cdot s_{\max})^2)$, \\ $\rho \sim \mathcal{U}([0.01, 0.20])$} \\
\midrule
\multicolumn{2}{c}{\textbf{\textit{Initial State Randomization}}} \\
Initial Body Velocity $(v_{\text{init}})$ & $\mathcal{U}([1.0, 2.5])$ m/s \\
Initial Body Lean Angle $(\varphi_{\text{init}})$ & $\mathcal{U}([-10, 10])$ deg \\
Initial Servo Motor Position $(\theta_{m,\text{init}})$ & $\mathcal{U}([-20, 20])$ deg \\
Initial Hub Motor Velocity $(\omega_{m, \text{init}})$ & $\mathcal{U}([0, 3.0])$ rad/s \\
\midrule  
\multicolumn{2}{c}{\textbf{\textit{Task / Command Randomization}}} \\
Target Velocity Command $(v_{\text{cmd}})$ & \makecell[l]{$v_{\text{cmd}} \sim \mathcal{U}([1.0, 5.0])$ m/s, \\ $t_{\text{resample}} \sim \mathcal{U}([3,5])$ s} \\
Target Heading Command $(\delta_{\text{cmd}})$ & \makecell[l]{$\delta_{\text{cmd}} \sim \mathcal{U}([-10, 10])$ deg, \\ $t_{\text{resample}} \sim \mathcal{U}([3,5])$ s} \\
\bottomrule  
\end{tabularx}  
\end{table}  

\subsubsection{Dynamics Randomization}

To address parametric uncertainties and modeling inaccuracies, we randomize key physical properties (see \Cref{tab:actual_domain_randomization}). Specifically, mass, center-of-mass height and terrains are varied to reflect configuration changes, while friction and actuator gains are randomized to capture diverse terrains and hardware variations. Additionally, Gaussian noise (scaled by signal magnitude $s_{\text{max}}$) is injected to simulate sensor imperfections, ensuring distributional robustness.

\subsubsection{Initial State Randomization}

We randomize the bicycle's initial kinematic states at the start of each episode. This forces the agent to learn policies that can recover from diverse unstable configurations (e.g., varying initial speeds and tilt angles) rather than relying on a static, upright start, resiliently ensuring safety during real-world engage-disengage transitions.

\subsubsection{Task and Command Randomization}

To develop a versatile controller, target velocity and heading commands are dynamically altered at random time intervals during training. This compels the policy to learn a generalizable state-command mapping rather than overfitting to a static equilibrium, ensuring reliable high-level directive tracking.
% enabling the physical agent to reliably track high-level navigation directives.

\subsection{System Architecture and Implementation}

\subsubsection{Physical Simulation Framework}

\begin{figure}[htbp]
    \centering
    \includegraphics[width=0.9\linewidth]{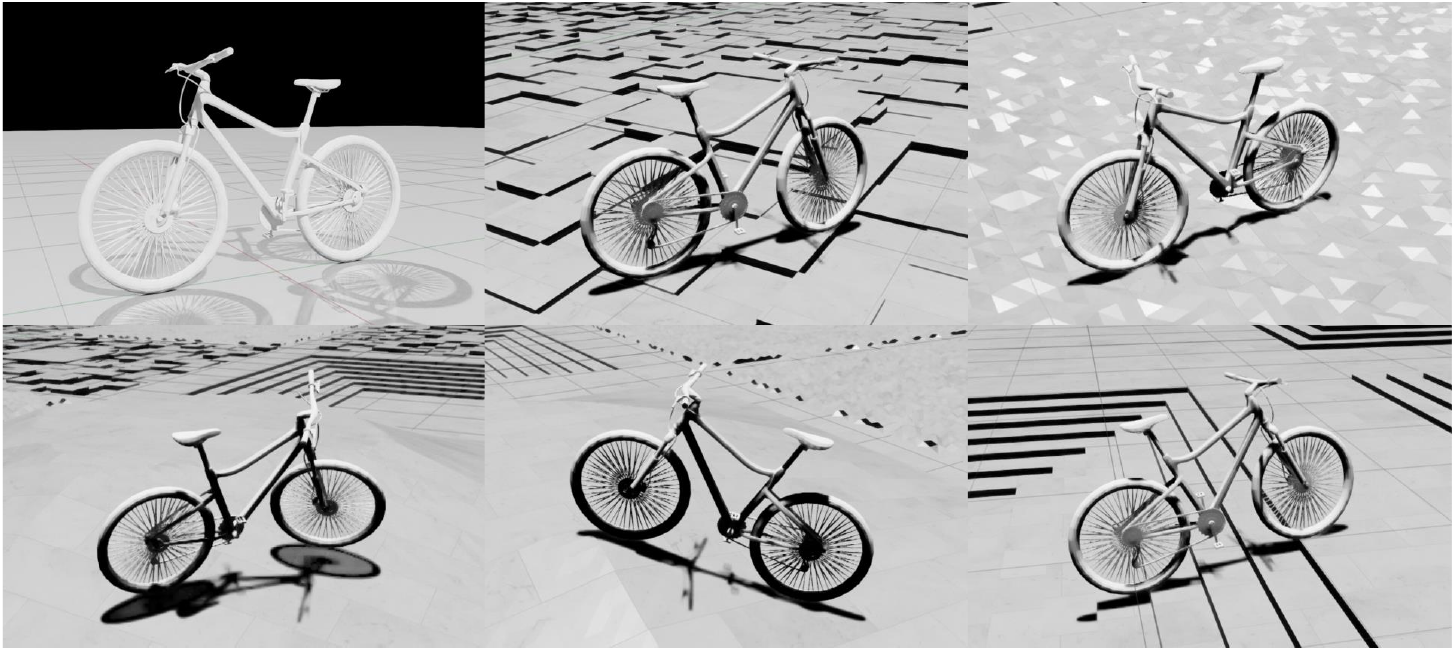}
    \caption{Bicycle and terrain modeling in Isaac Sim.}
    \label{fig:isaacsim}
\end{figure}

The training environment utilizes NVIDIA Isaac Sim for GPU-accelerated, parallel physics simulation. It explicitly incorporates advanced contact and actuator dynamics, variable surface friction, and realistic IMU sensor properties. This comprehensive modeling exposes the policy to complex real-world vehicle-terrain nonlinearities, as illustrated in \Cref{fig:isaacsim}. Detailed vehicle kinematics and dynamic parameters are provided in our supplementary \href{https://cpnt-lab.github.io/CycleRL/}{video}.

% 这张图是否有必要保留，详细阐述如何建模是否会更好，
% 例如车体由哪几部分构成，包含哪些关节

\subsubsection{Hierarchical Control Architecture}

To support real-time onboard deployment, we designed a hierarchical architecture consisting of input processing, policy inference, and actuator control layers. An asynchronous multi-process framework efficiently manages the high-frequency data flow, ensuring minimal end-to-end latency for the 50 Hz control loop. Further details regarding the software stack and signal flow are available in the supplementary \href{https://cpnt-lab.github.io/CycleRL/}{video}.

\section{Experiments}
\label{sec:eval}
\subsection{Evaluation Metrics and Implementation Details}

\subsubsection{Metrics Definition on Performance}

% We establish a comprehensive evaluation framework encompassing multiple performance dimensions critical for autonomous bicycle control systems. \Cref{tab:performance_metrics} summarizes the evaluation criteria across three categories.
We establish a comprehensive evaluation framework assessing three critical dimensions for autonomous bicycle control, as stated in \Cref{tab:performance_metrics}.

\begin{table}[htbp]  
    \centering  
    \caption{Evaluation Metrics on Performance}  
    \label{tab:performance_metrics}  
    \begin{tabularx}{0.99\columnwidth}{l X} % l: left-aligned, X: auto-wrapping text  
        \toprule
        \textbf{Metric} & \textbf{Definition} \\
        \midrule
        \multicolumn{2}{c}{\textbf{\textit{Balance Metrics}}} \\
        \multirow{2}{*}{Balance Success Rate (BSR)} & Percentage of episodes maintaining $|\varphi|<0.5$ rad. \\
        \multirow{2}{*}{Balance Recovery Time (BRT)} & Recovery time from disturbances exceeding 0.3 rad. \\   
        \multirow{2}{*}{Maximum Balance Duration (MBD)} & Longest continuous balancing period achieved. \\   
        \multirow{2}{*}{Critical Angle Tolerance (CAT)} & Maximum recoverable roll angle maintaining BSR $>$ 95\%. \\   
        \midrule
        \multicolumn{2}{c}{\textbf{\textit{Control Metrics}}} \\
        \multirow{2}{*}{Heading Tracking Error (HTE)} & RMS error between commanded and actual heading. \\   
        \multirow{2}{*}{Velocity Tracking Error (VTE)} & Mean absolute error in longitudinal velocity control. \\   
        \multirow{2}{*}{System Response Latency (SRL)} & Time to reach within 10\% of the target command value. \\   
        \midrule
        \multicolumn{2}{c}{\textbf{\textit{Robustness Metrics}}} \\
        \multirow{2}{*}{Maximum Noise Tolerance (MNT)} & Highest sensor noise level maintaining BSR $>$ 95\%. \\   
        \multirow{2}{*}{Minimum Sustaining Speed (MSS)} & Minimum speed maintaining BSR $>$ 95\%. \\   
        \bottomrule
    \end{tabularx}
\end{table}  

\subsubsection{Training Configuration and Hyperparameters}
Training details, including configurations and hyperparameters, are provided in supplementary \href{https://cpnt-lab.github.io/CycleRL/}{video}. Our training infrastructure leverages NVIDIA Isaac Sim, achieving over 700,000 simulation steps per second across 16,384 parallel environments on NVIDIA RTX 4090, ensuring a considerable iteration speed.

\subsection{Simulation Experiments and Quantitative Analysis}

\subsubsection{Evaluation on Simulation Performance}
\label{sec:baseline}

We conduct comprehensive experiments across 10,000 distinct episodes to establish core performance characteristics. Evaluations are performed under ideal dynamics (i.e., no physical parameter randomization) but retain initial state and command randomization, with target velocities sampled from $\mathcal{U}([2.0, 3.0])$ m/s for our policy and fixed at 2.0 m/s for baselines.

To validate the proposed approach, we benchmark against calibrated PID and LQR baselines, which serve as canonical benchmarks for self-balancing systems and, comparable to CycleRL, require only limited system identification. PID baseline features a dual-loop structure (steer/balance) with velocity-dependent gain scaling and base gains were tuned to $K_p=4.0$ and $K_d=0.4$ via Ziegler-Nichols at 2.0 m/s. LQR baseline utilizes real-time Riccati equation solving on a linearized model, whose cost weights were set to $Q=\text{diag}[20.0, 6.0, 3.5]$ and $R=1.5$, with heading tracking achieved by mapping $\delta_\text{cmd}$ to an equilibrium roll angle. We also implement the advanced Modified Linear Control Law (MLCL)~\cite{xiong2024steering} for a stronger and comparable model-based evaluation. \Cref{tab:simulation_comparison} presents the quantitative results across our evaluation metrics.

\begin{table}[ht]
    \centering
    \caption{Comparative Simulation Performance}
    \label{tab:simulation_comparison}
    % 使用 resizebox 强制将表格压缩到单栏宽度
    \resizebox{\columnwidth}{!}{
        \begin{tabular}{lcccc}
            \toprule
            \textbf{Metric $^1$} \tnote{1} & \textbf{PID} & \textbf{LQR} & \textbf{MLCL} & \textbf{Ours} \\
            \midrule
            \multicolumn{5}{c}{\textbf{\textit{Balance Performance}}} \\
            Success Rate (\%) $\uparrow$ & 76.50 & 93.24 & 96.43 & \textbf{99.90} \\
            Recovery Time (s) $\downarrow$ & 1.41\tiny{$\pm$0.22} & 1.47\tiny{$\pm$0.21} & 1.21\tiny{$\pm$0.04} & \textbf{1.05\tiny{$\pm$0.18}} \\
            Max. Duration (s) $\uparrow$ & 15.0 & 148.9 & 603.2 & \textbf{$>$1,800} \\
            Crit. Angle Tol. ($^\circ$) $^2$ $\uparrow$ & 22.24 & 21.40 & 24.65 & \textbf{27.79} \\
            \midrule
            \multicolumn{5}{c}{\textbf{\textit{Control Performance}}} \\
            Heading Error ($^\circ$) $\downarrow$ & 4.04\tiny{$\pm$1.51} & 1.82\tiny{$\pm$0.64} & 2.87\tiny{$\pm$1.66} & \textbf{1.15\tiny{$\pm$0.72}} \\
            Velocity Error (m/s) $^3$ $\downarrow$ & $\dagger$ & $\dagger$ & $\dagger$ & \textbf{0.18\tiny{$\pm$0.26}} \\
            Response Latency (s) $\downarrow$ & 1.67\tiny{$\pm$1.12} & 1.27\tiny{$\pm$0.61} & 1.46\tiny{$\pm$1.10} & \textbf{1.06\tiny{$\pm$0.17}} \\
            \midrule
            \multicolumn{5}{c}{\textbf{\textit{Robustness Indicators}}} \\
            Max. Noise Tol. $^4$ { }$\uparrow$ & $\dagger$ & $\dagger$ & $\dagger$ & \textbf{0.32} \\
            Min. Sustaining Spd (m/s) $\downarrow$ & 1.84 & 1.71 & 1.50 & \textbf{1.05} \\
            \bottomrule
        \end{tabular}
    }
    \begin{tablenotes}
        \leftskip=-0.6em
        % \scriptsize
        \fontsize{6.9pt}{8pt}\selectfont
        \item[1] $^1$ Reported as Mean $\pm$ Standard Deviation where applicable; $\dagger$ means not applicable.
        \item[2] $^2$ Reported as the minimum of both sides (conservative estimate).
        \item[3] $^3$ Baselines are tested primarily for lateral balance at constant velocities.
        \item[4] $^4$ MNT represents the max tolerable noise amplitude fraction.
    \end{tablenotes}
\end{table}

The results confirm our controller's superior performance compared to traditional methods. CycleRL achieves a 99.90\% BSR, significantly outperforming the PID (76.50\% BSR), LQR (93.24\% BSR) and MLCL (96.43\% BSR) baselines. Remarkably, CycleRL recovers faster from perturbations (1.05 s) than PID, LQR, and MLCL, while tolerating extreme roll angles up to 27.79° (vs. 24.65° for MLCL). Furthermore, CycleRL maintains robust stability at speeds as low as 1.05 m/s, where classical baselines fail, while simultaneously ensuring precise heading tracking (1.15$^\circ$ HTE).

% {\color{red}As indicated in \Cref{tab:simulation_comparison}, CycleRL exhibits superior performance across all metrics. It achieves considerable stability (99.90\% BSR) and recovers faster from perturbations (1.05 s) than PID, LQR, and MLCL. Furthermore, it tolerates extreme roll angles up to 27.79$^\circ$ and maintains robust stability at speeds as low as 1.05 m/s, where classical baselines fail, while simultaneously ensuring precise heading tracking (1.15$^\circ$ HTE) and velocity tracking (0.18 m/s VTE).}

\subsection{Convergence Analysis and Learning Dynamics}

\begin{figure}[htbp]
    \centering
    \begin{subfigure}[b]{0.48\textwidth}
        \centering
        \includegraphics[width=0.63\linewidth]{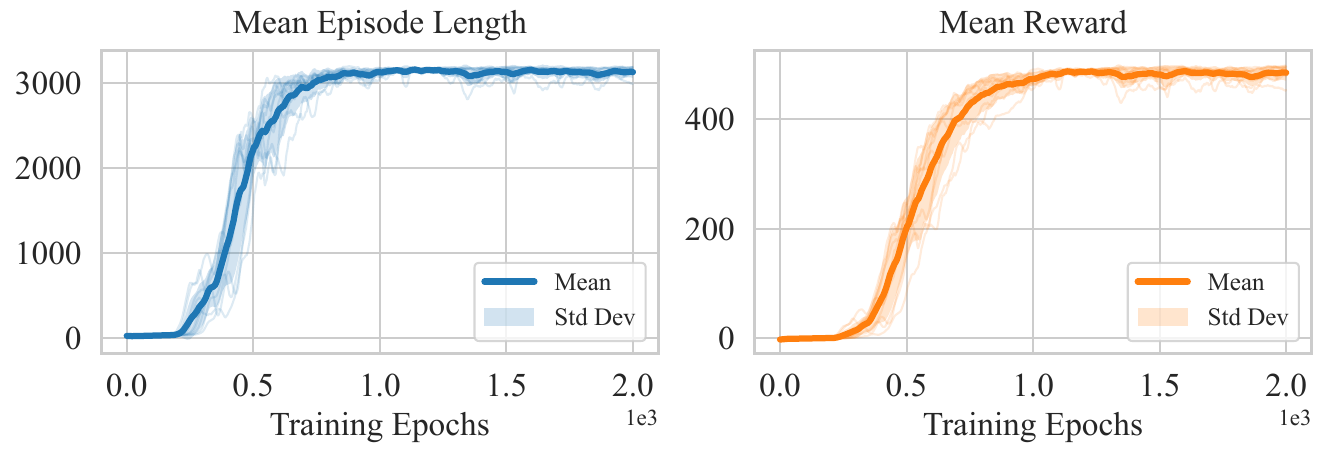}
        \caption{Mean Episode Length \& Mean Total Reward}
        \label{fig:mean_rewards}
    \end{subfigure}
    % \hfill
    \vspace{0.8em}
    
    \begin{subfigure}[b]{0.48\textwidth}
        \centering
        \includegraphics[width=0.95\linewidth]{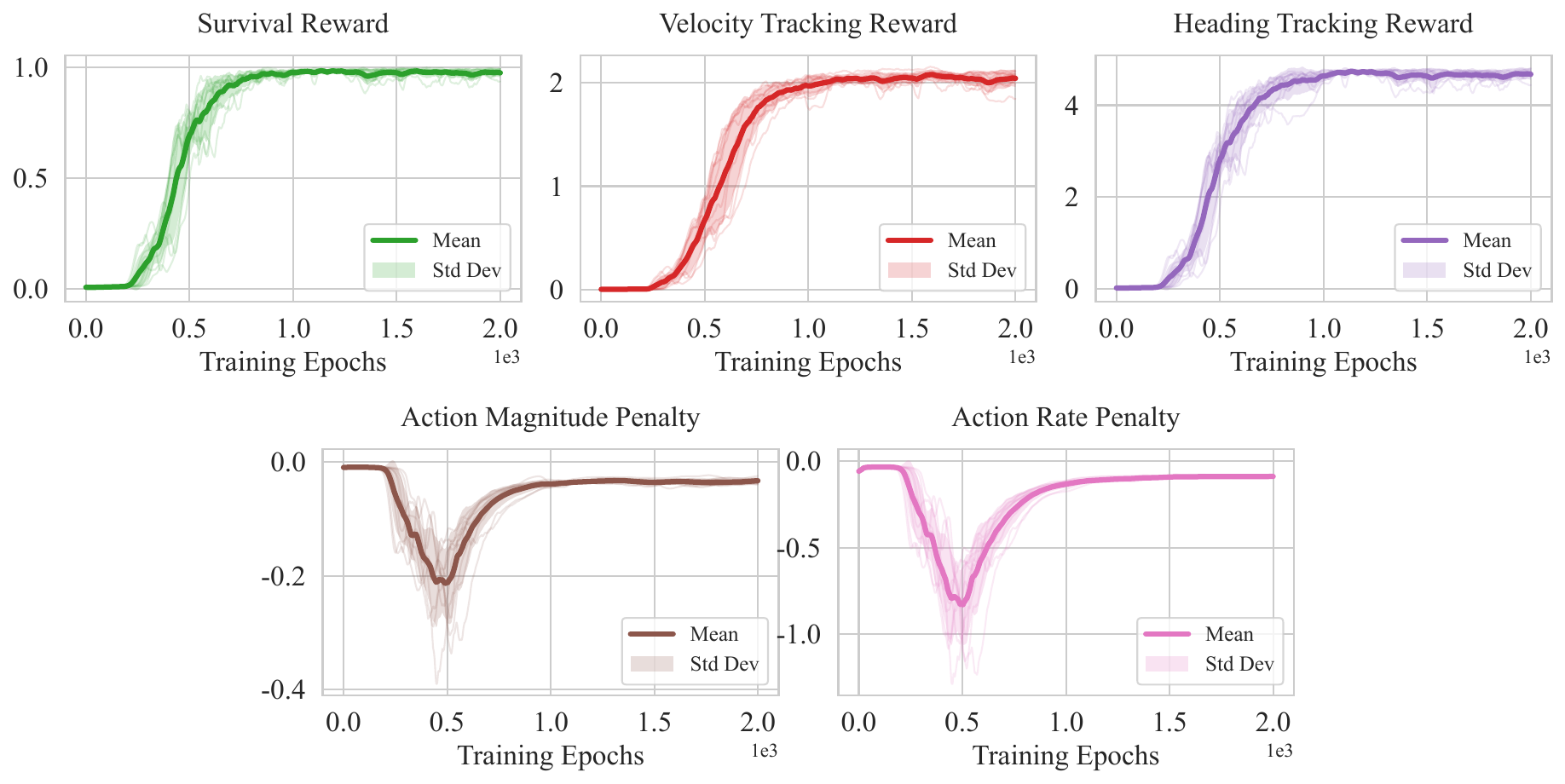}
        \caption{Training Rewards Curves}
        \label{fig:reward_components}
    \end{subfigure}

    \caption{Training curves and convergence analysis.}
    \label{fig:convergence}
\end{figure}

\Cref{fig:convergence} illustrates the policy's training progression under 10 random seeds. Stable convergence is achieved after approximately 1,000 training epochs (700M simulation steps). As shown in \Cref{fig:mean_rewards}, both Mean Episode Length and Mean Total Reward exhibit a monotonic increase and eventual plateau. %The agent consistently learned to prolong its operational duration and maximize cumulative rewards.

% The learning dynamics, particularly the penalty curves in \Cref{fig:reward_components}, reveal a natural two-phase curriculum learning effect. An initial phase of aggressive control, marked by a dip in the penalty values, allows the agent to quickly master basic stability. This is followed by a refinement phase, where the penalties decrease (rise toward zero) as the agent learns to perform the task with smoother, more efficient actions.

The learning dynamics, particularly the penalty curves in \Cref{fig:reward_components}, reveal an implicit curriculum learning effect: an initial phase of aggressive control (marked by penalty dips) allows the agent to quickly master basic stability, followed by a refinement phase where penalties rise toward zero as the policy learns smoother and more efficient actions.

\subsection{Robustness Verification Experiments}

To assess deployment readiness, we evaluate robustness by individually introducing velocity variations, sensor degradation, and terrain disturbances to the baseline setup (\Cref{sec:baseline}) under 10,000 trials. \Cref{tab:robustness_results} summarizes the per-formance under various conditions, where the ``No noise'' and ``Flat plane'' cases replicate the ideal baseline conditions.

% 这张图是否有必要保留，信息量比较低
% \begin{figure}[htbp]
%     \centering
%     \includegraphics[width=0.95\linewidth]{figs/robustness_evaluation_barchart.pdf}
%     \caption{Robustness evaluation results.}
%     \label{fig:robustness}
% \end{figure}

The evaluation confirms the system's performance and resilience. Considerable performance is achieved on ideal sensor and terrain configurations at low speeds (99.49\% BSR) and at higher speeds (86.21\% BSR). The system effectively handles sensor degradation, maintaining a BSR of 96.38\% with severe noise and 90.84\% with intermittent dropouts. On complex terrains, the system exhibits a graceful performance degradation ranging from 7.85\% on slopes to 23.30\% over gentle steps, compared to the ideal condition. Balance Recovery Time (BRT) and tracking errors (HTE, VTE) also remain well-maintained across diverse conditions. % Overall, these results validate the system's considerable robustness for operation.

Notably, a decline in BSR is observed at higher velocities. Although a bicycle exhibits enhanced passive self-stability during high-speed straight-line driving, this degradation stems from the compounding complexities of active high-speed maneuvering. Under our evaluation protocol featuring randomized velocity and heading commands, the system is potentially forced into sharp turns at high speeds. In this regime, lateral acceleration scales quadratically with velocity ($a_y \propto v^2\delta$), rendering the closed-loop system hypersensitive to steering inputs and actuator phase lag, and severely intensifying the multi-objective coupling between heading tracking and roll stability. Moreover, the fixed 50~Hz control bandwidth, coupled with the deficiencies of the PhysX-based Isaac Sim environment in simulating high-speed physical dynamics and collisions, further exacerbates the brittleness of high-speed cornering.

\subsection{Ablation Studies}
\label{sec:ablation}

\subsubsection{Ablation Study of Reward Functions}

We systematically investigate individual reward component contributions through controlled ablation experiments. \Cref{tab:reward_ablation} presents performance degradation when each component is removed.
%  and \Cref{fig:reward_ablation}

\begin{table}[t]
    \centering
    \caption{Ablation Results of Reward Functions}
    \label{tab:reward_ablation}
    \begin{tabular}{lcccc}
        \toprule
        \textbf{Configuration} & \textbf{BSR (\%) $\uparrow$} & \textbf{HTE (deg) $\downarrow$} & \textbf{VTE (m/s) $\downarrow$} \\
        \midrule
        Complete Reward & 99.90 & \textbf{1.15} & \textbf{0.18} \\
        No Survival Reward & \textbf{99.99} \textcolor{blue}{(+0.09)} & 1.27 \textcolor{red}{(+0.12)} & 0.19 \textcolor{red}{(+0.01)} \\
        No Velocity Tracking & 96.71 \textcolor{red}{(-3.19)} & \textbf{1.15} \textcolor{blue}{(-0.00)} & 0.59 \textcolor{red}{(+0.41)} \\
        No Heading Tracking & 99.87 \textcolor{red}{(-0.03)} & 6.30 \textcolor{red}{(+5.15)} & 0.21 \textcolor{red}{(+0.03)} \\
        No Action Penalty & \textbf{99.99} \textcolor{blue}{(+0.09)} & 1.72 \textcolor{red}{(+0.57)} & 0.38 \textcolor{red}{(+0.20)} \\
        No Rate Penalty & 99.87 \textcolor{red}{(-0.03)} & 2.29 \textcolor{red}{(+1.14)} & 0.55 \textcolor{red}{(+0.37)} \\
        \bottomrule
    \end{tabular}
\end{table}

% \begin{figure}[htbp]
%     \centering
%     \includegraphics[width=0.95\linewidth]{figs/reward_ablation_barchart_v3.pdf}
%     \caption{Results of Reward Ablation.}
%     \label{fig:reward_ablation}
% \end{figure}

The ablation analysis reveals that tracking rewards are crucial. Removing the velocity tracking reward leads to a significant stability performance drop (BSR -3.19\%, VTE +0.41 m/s); the heading tracking reward follows a similar pattern. Interestingly, removing the survival reward and action penalty slightly improves the BSR, suggesting a potential conservative bias. Finally, the absence of the rate penalty harms control smoothness, increasing both HTE and VTE.

\begin{table*}[htbp]
    \centering
    \caption{Robustness Evaluation}
    \label{tab:robustness_results}
    \begin{tabular}{lcccccc}
        \toprule
        \textbf{Test Category} & \textbf{Condition} & \textbf{BSR (\%) $\uparrow$} & \textbf{BRT (s) $\downarrow$} & \textbf{HTE (deg) $\downarrow$} & \textbf{VTE (m/s) $\downarrow$} & \textbf{Notes} \\
        \midrule
        \multirow{2}{*}{Velocity Range} % & Ultra-low (0.1-1.0 m/s) & 68.4 & 4.2 & 3.8 & 0.15 & Challenging stability regime \\
        & 1.0-3.0 m/s & 99.49 & 1.58 & 1.57 & 0.23 & Low speed \\
        & 3.0-5.0 m/s & 86.21 & 2.05 & 1.56 & 0.28 & High speed \\
        \midrule
        \multirow{3}{*}{Sensor Degradation} & No noise & \textbf{99.90} & \textbf{1.05} & \textbf{1.15} & \textbf{0.18} & Baseline performance \\
        & Maximum noise level & 96.38 & 2.90 & 3.46 & 0.78 & 20\% amplitude noise \\
        % & 5× noise level & 84.1 & 4.1 & 3.7 & 0.15 & Significant noise impact \\
        & Intermittent dropouts & 90.84 & 1.62 & 1.92 & 0.48 & 10\% data loss rate \\
        \midrule
        \multirow{5}{*}{Terrain Type} & Flat plane & \textbf{99.90} & \textbf{1.05} & \textbf{1.15} & \textbf{0.18} & Ideal conditions \\
        & Rough pavement & 88.47 & 1.66 & 1.44 & \textbf{0.18} & Standard road surface \\
        & Gravel surface & 85.87 & 2.15 & 1.69 & 0.21 & Reduced traction \\
        & Slope terrain & 92.05 & 1.74 & 1.35 & 0.23 & Slope adaptation \\
        & Gentle steps & 76.60 & 1.80 & 1.31 & 0.22 & Complex terrain \\
        \bottomrule
        \end{tabular}
\end{table*}

% \subsubsection{State Space Representation Analysis}

% \Cref{tab:state_ablation} compares alternative state representations to validate our 8-dimensional observation space design.

% \begin{table}[ht]
%     \centering
%     \caption{State Space Representation Ablation Results}
%     \label{tab:state_ablation}
%     \begin{tabular}{lcccc}
%         \hline
%         \textbf{State Configuration} & \textbf{Dimensions} & \textbf{BSR (\%)} & \textbf{Training Time (h)} & \textbf{SRL (ms)} \\
%         \hline
%         Baseline (8D) & 8 & 94.7 & 12.0 & 12.5 \\
%         Minimal (4D) & 4 & 78.3 & 8.2 & 9.1 \\
%         Extended (12D) & 12 & 95.8 & 18.7 & 16.3 \\
%         History-Augmented (16D) & 16 & 96.1 & 24.5 & 21.8 \\
%         Filtered Observations & 8 & 91.2 & 14.3 & 15.7 \\
%         \hline
% \end{tabular}
% \end{table}

% The analysis demonstrates that our 8-dimensional state space achieves optimal balance between performance and computational efficiency. Minimal representations sacrifice performance significantly, while extended representations provide marginal improvements at substantial computational cost.

\subsubsection{Impact Assessment of Domain Randomization}

\Cref{tab:domain_ablation} evaluates the contribution of different randomization strategies.

\begin{table}[ht]
    \centering
    \caption{Component Analysis of Domain Randomization}
    \label{tab:domain_ablation}
    \begin{tabular}{lcccc}
        \toprule
        \makecell[l]{\textbf{Randomization}\\\textbf{Strategy}} & \textbf{BSR (\%) $\uparrow$} & \textbf{HTE (deg) $\downarrow$} & \textbf{VTE (m/s) $\downarrow$} \\
        \midrule
        Full Randomization & 99.90 & \textbf{1.15} & \textbf{0.18} \\
        No Randomization & \textbf{99.99} \textcolor{blue}{(+0.09)} & 5.16 \textcolor{red}{(+4.01)} & 0.49 \textcolor{red}{(+0.31)} \\
        Dynamics-Only & 99.95 \textcolor{blue}{(+0.05)} & 2.49 \textcolor{red}{(+1.34)} & 0.43 \textcolor{red}{(+0.25)} \\
        Initial States-Only & \textbf{99.99} \textcolor{blue}{(+0.09)} & 4.03 \textcolor{red}{(+2.88)} & 0.57 \textcolor{red}{(+0.39)} \\
        Command-Only & 99.19 \textcolor{red}{(-0.71)} & 1.72 \textcolor{red}{(+0.57)} & 0.20 \textcolor{red}{(+0.02)} \\
        Terrain-Only & \textbf{99.99} \textcolor{blue}{(+0.09)} & 3.84 \textcolor{red}{(+2.69)} & 0.58 \textcolor{red}{(+0.40)} \\
        % Progressive Randomization & 94.2 & 85.9 & 8.3 \\
        \bottomrule
    \end{tabular}
\end{table}

The analysis reveals the significant impact of different randomization strategies. The Full Randomization strategy achieves the best overall performance with the lowest tracking errors (HTE 1.15 deg, VTE 0.18 m/s). Single-component randomizations, e.g., Dynamics-Only, Initial States-Only or Terrain-Only, also result in substantial performance loss, particularly in heading and velocity control. 

Interestingly, the ``No Randomization'' strategy yields a marginally higher BSR (99.99\%) in ideal simulation conditions compared to the full strategy, indicating a performance-robustness trade-off. The extensive randomization required for sim-to-real transfer inherently introduces conservatism into the policy, slightly sub-optimizing peak theoretical performance to guarantee broad adaptability. However, as noted by the severe degradation in precision (HTE +4.01°, VTE +0.31 m/s) when deploying without full randomization, this slight conservative bias is a necessary and acceptable compromise for safe and effective deployment across varying real-world conditions.
% These results confirm that a comprehensive, full randomization is essential for robust sim-to-real transfer.

\subsubsection{Sensitivity Analysis of Reward Weights}

To validate our reward configuration, we conducted a local sensitivity analy-sis~\cite{xiong2024steering} by independently scaling each weight ($\lambda_{\text{surv}}$, $\lambda_{\text{vel}}$, $\lambda_{\text{head}}$, $\lambda_{\text{act}}$, $\lambda_{\text{rate}}$) from 0.1$\times$ to 5.0$\times$ of its nominal value.

\begin{figure}[htbp]
    \centering
    \includegraphics[width=0.98\linewidth]{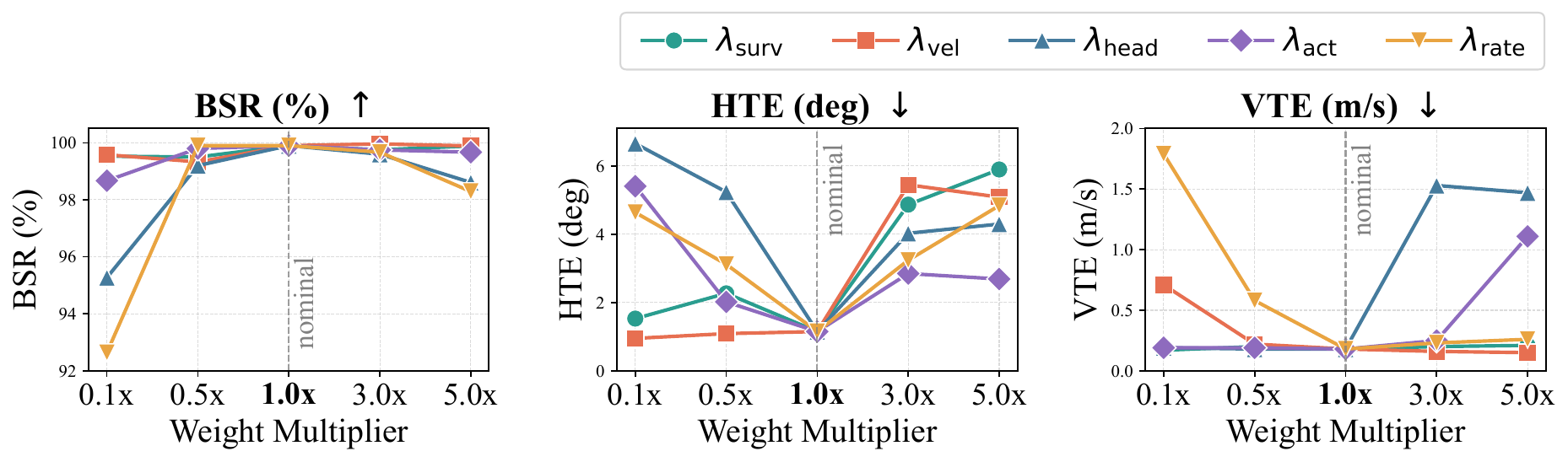}
    \caption{Local Sensitivity Analysis of Reward Weights.}
    \label{fig:reward_sensitivity}
\end{figure}

As illustrated in \Cref{fig:reward_sensitivity}, the nominal configuration (1.0$\times$) generally achieves the most balanced performance across BSR, HTE, and VTE. Deviating from these values degrades overall performance; for instance, overly prioritizing velocity tracking notably increases heading errors, while insufficient action rate penalties compromise balancing stability. Conversely, varying the survival reward minimally impacts final performance, as it primarily accelerates and stabilizes training. Overall, these findings validate our weight configuration as an empirically robust operating point. While this local analysis avoids computationally prohibitive exhaustive searches, future work could employ Bayesian optimization or joint random search to efficiently capture inter-parameter dependencies.

% To validate our reward configuration, we independently scaled each weight ($\lambda_{vel}$, $\lambda_{head}$, $\lambda_{act}$, $\lambda_{rate}$) from 0.1$\times$ to 5.0$\times$ of its nominal value. As illustrated in Fig. X, the nominal configuration (1.0$\times$) consistently achieves the most balanced performance across BSR, HTE, and VTE. Deviating from these values degrades overall performance; for instance, overly prioritizing velocity tracking notably increases heading errors, while insufficient action rate penalties compromise balancing stability. These results substantiate that the chosen weights are empirically robust and represent an optimal operating point, effectively addressing the heuristic nature of reward shaping.

\subsection{Implementation and Validation of Hardware System}

\subsubsection{Construction of Physical Platform}

\begin{figure}[htbp]
    \centering
    \includegraphics[width=0.8\linewidth]{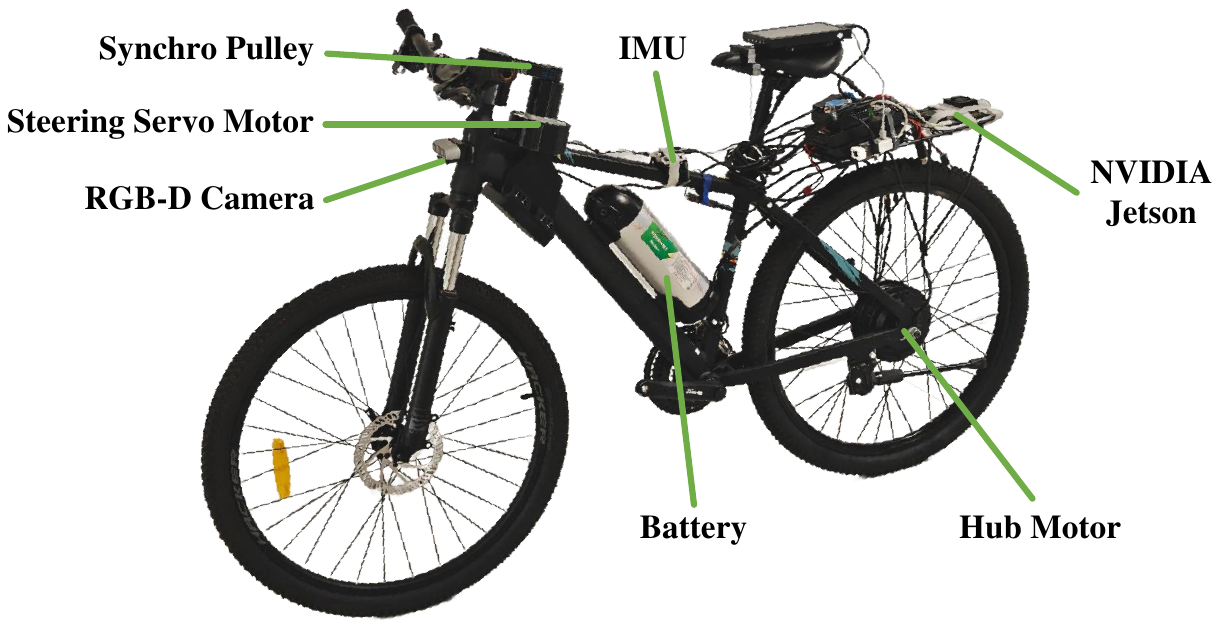}
    \caption{Construction of hardware platform.}
    \label{fig:hardware_construction}
\end{figure}

Our hardware validation utilizes a custom-designed bicycle platform, tailored for real-world evaluation, as shown in \Cref{fig:hardware_construction}. Specifications of this hardware are detailed in \Cref{tab:hardware_specs}.

\begin{table}[ht]
    \centering
    \caption{Specifications of Hardware Platform}
    \label{tab:hardware_specs}
    \begin{tabular}{lc}
        \toprule
        \textbf{Component} & \textbf{Specification} \\
        \midrule
        \multicolumn{2}{c}{\textbf{\textit{Mechanical Platform}}} \\
        % Base vehicle & Modified Mountain Bike \\
        Wheelbase & 1,100 mm \\
        Total mass & 25.0 kg (including electronics) \\
        Center of gravity height & 0.65 m \\
        % Wheel diameter & 26 inches \\
        \midrule
        \multicolumn{2}{c}{\textbf{\textit{Computational System}}} \\
        Main processor & Jetson Orin NX 16GB \\
        % Memory & 16 GB LPDDR5 \\
        % Storage & 512 GB NVMe SSD \\
        Operating system & Ubuntu 20.04 \\
        \midrule
        \multicolumn{2}{c}{\textbf{\textit{Sensing and Actuation}}} \\
        IMU & WheelTec N100 \\
        Steering servo motor & Unitree GO-M8010-6 \\
        Hub motor & BaFang RM G020.500.D \\
        Power system & 48V 9.6Ah lithium battery \\
        \bottomrule
    \end{tabular}
\end{table}

The mechanical design emphasizes modularity and reliability, utilizing off-the-shelf components to ensure reproducibility. Computationally, the NVIDIA Jetson board provides sufficient processing power for real-time neural network inference while maintaining power efficiency for extended operation.

\subsubsection{Validation of Real-World Deployment}

To evaluate the policy's real-world transferability, \Cref{tab:realworld_results} systematically compares its performance between simulation and real world.

\begin{table}[ht]
    \centering
    \caption{Comparison of Sim and Real Performance}
    \label{tab:realworld_results}
    \begin{tabular}{lccc}
        \toprule
        \textbf{Metric} & \textbf{Sim} & \textbf{Real} & \makecell{\textbf{Transfer}\\\textbf{Ratio} $\uparrow$} \\
        \midrule
        Balance Success Rate (\%) & 99.90 & 95.00 & 0.95 \\
        Balance Recovery Time (s) & 1.05 & 1.29 & 0.81 \\
        Maximum Balance Duration (s) & $>$1,800 & $>$1,800 & 1.00 \\
        Critical Angle Tolerance (deg) & 27.79 & 20.12 & 0.72 \\
        Heading Tracking Error (deg) & 1.15 & 2.04 & 0.56 \\
        Velocity Tracking Error (m/s) & 0.18 & 0.01 & 1.00 \\
        System Response Latency (s) & 1.06 & 1.10 & 0.96 \\
        Minimum Sustaining Speed (m/s) & 1.05 & 1.33 & 0.79 \\
        % Maximum Noise Tolerance & 0.32 & 0.02 & 0.75 \\
        % Terrain Adaptability Score (\%) & 0.87 & 0.05 & 0.75 \\
        % Payload Sensitivity Index (kg) & 10 & 0.08 & 0.75 \\
        \bottomrule
    \end{tabular}
\end{table}

% \begin{figure}[htbp]
%     \centering
%     \includegraphics[width=0.95\linewidth]{figs/sim_vs_real_normalized_barchart.pdf}
%     \caption{Transference analysis of real-world deployment.}
%     \label{fig:realworld_results}
% \end{figure}

% \begin{table}[ht]   % 实车上RL与baseline的对比，可加可不加，如果实车调不出来就不加
%     \centering
%     \caption{Real-World Performance Comparison}
%     \label{tab:real_comparison}
%     \begin{tabular}{lccc}
%         \toprule
%         \textbf{Metric} & \textbf{LQR} & \textbf{CycleRL} & \textbf{Improvement} \\
%         \midrule
%         Min. Sustaining Speed (m/s) $\downarrow$ & 2.50 & \textbf{1.33} & +46.8\% \\
%         Max. Balance Duration (s) $\uparrow$ & 120 & \textbf{$>$1,800} & $>$15x \\
%         Recovery Time (s) $\downarrow$ & 3.5 & \textbf{2.29} & +34.5\% \\
%         Steering Jitter (deg/s) $\downarrow$ & 5.4 & \textbf{2.1} & (Smoother) \\
%         \bottomrule
%     \end{tabular}
% \end{table}

The results confirm an effective sim-to-real transfer, achieving transfer ratios from 0.56 to 1.00 across key metrics, where a value of 1.00 indicates perfect transference. The most notable degradation, Heading Tracking Error (ratio: 0.56), is attributed to unmodeled IMU noise. Conversely, Velocity Tracking Error was nearly eliminated (ratio: 1.00), as the velocity command was directly mapped to motor speed, bypassing inference. These results validate the policy's real-world effectiveness.

% Notably, real-world comparisons with the baselines are omitted, as they failed to directly stabilize the physical platform due to severe sensitivity to model mismatches. This stark contrast underscores that our DRL approach, combined with domain randomization, effectively bridges the Sim-to-Real gap that rendered traditional methods inoperable.

Notably, real-world baseline comparisons were omitted since they failed to directly stabilize the physical platform. Under zero-shot transfer constraint, baselines were tuned solely in simulation and deployed without hardware fine-tuning, but their reliance on nominal linearized models made them highly sensitive to sim-to-real mismatches. In contrast, our domain-randomized DRL policy learned robustness to such uncertainties and successfully transferred to hardware.

To validate the policy's efficacy as a robust controller, we employed two command modalities. In the \textit{human-in-the-loop} configuration, high-level directives ($v_{\text{cmd}}, \delta_{\text{cmd}}$) were provided by a remote operator, with the CycleRL policy autonomously managing the actuation for stability and steering. The policy was tested across diverse conditions, including asphalt roads, gravel paths, rugged grass, ramps, varying payloads and velocities, tire anomalies and lateral perturbations. To further demonstrate higher-level autonomy, we integrated a \textit{vision-based lane tracking system}~\cite{qin2020ultra}, in which RGB images from a front-mounted Intel RealSense D435i camera are used for lane detection and the deviation between the detected lane center and the vehicle's heading is fed into a tuned PID controller to yield the reference steering command. Furthermore, we validate the framework's generalizability via successful zero-shot transfer to a scooter platform using the identical policy, with all deployment videos available at \href{https://cpnt-lab.github.io/CycleRL/}{video link}.

% To further validate generalizability, we deployed the identical policy to a scooter platform via zero-shot transfer, achieving considerable stability without any re-training or fine-tuning. Videos of both command modalities and cross-platform validation are available at \href{https://cpnt-lab.github.io/CycleRL/}{video link}.

% The results confirm a successful sim-to-real transfer, achieving transfer ratios between 0.67 and 0.92 across key performance metrics. Despite unmodeled dynamics and environmental noise, the physical platform attained a Balance Success Rate (BSR) of 87.2\% and demonstrated continuous outdoor operation for 38.7 minutes. The system's robustness is further highlighted by its ability to recover from induced roll disturbances up to 30.1°.

% Additional real-world tests demonstrated robust rejection of lateral disturbances up to 35N across various terrains, including asphalt and grass. These comprehensive experimental results validate the practical viability of our deep reinforcement learning approach, proving its effectiveness and resilience for autonomous bicycle control in complex, real-world conditions.

\section{Conclusion}
\label{sec:conclusion}

This work presented a deep reinforcement learning framework for the robust lateral control of an autonomous bicycle. By training a direct perception-to-action policy, our model-free approach successfully navigates the vehicle's complex, underactuated dynamics, offering a promising alternative to traditional model-based controllers without requiring explicit analytical dynamic modeling and system identification.

Our key contributions include a composite reward function that balances stability with tracking objectives, and a comprehensive domain randomization strategy, which proved crucial for bridging the reality gap, enabling successful transfer of the learned policy. Extensive real-world experiments validated the controller's robustness across diverse conditions, confirming its viability for autonomous cycling applications.

While our method bypasses explicit dynamic modeling, it inherently shifts the engineering effort toward simulator construction, reward shaping, and hyperparameter tuning. To address this, future work will explore automated methods to reduce these heuristic dependencies, alongside extending the framework to encompass vision-based navigation and multi-agent coordination. Ultimately, the success of this approach highlights the potential of model-free DRL in advancing intelligent mobility for sustainable urban transportation.
\\

% {\appendix[Proof of the Zonklar Equations]
% Use $\backslash${\tt{appendix}} if you have a single appendix:
% Do not use $\backslash${\tt{section}} anymore after $\backslash${\tt{appendix}}, only $\backslash${\tt{section*}}.
% If you have multiple appendixes use $\backslash${\tt{appendices}} then use $\backslash${\tt{section}} to start each appendix.
% You must declare a $\backslash${\tt{section}} before using any $\backslash${\tt{subsection}} or using $\backslash${\tt{label}} ($\backslash${\tt{appendices}} by itself
%  starts a section numbered zero.)}


% You can manually copy in the resultant .bbl file and set second argument of $\backslash${\tt{begin}} to the number of references
%  (used to reserve space for the reference number labels box).

\bibliographystyle{IEEEtran}
\bibliography{refs}

% % \vspace{11pt}
% % \bf{If you include a photo:}\vspace{-33pt}
% \begin{IEEEbiography}[{\includegraphics[width=1in,height=1.25in,clip,keepaspectratio]{fig1}}]{Gelu Liu}
% is currently pursuing the M.S. degree in Information and Communication Engineering at the School of Electronics and Communication Engineering, Sun Yat-sen University, China. He received the Bachelor of Engineering degree from Sun Yat-sen University, China, in 2022. His research interests include intelligent and autonomous navigation technologies of unmanned platforms in unknown environments based on deep reinforcement learning (DRL) and efficient spatial encoding methods.
% \end{IEEEbiography}



\vfill

\end{document}